\documentclass[letterpaper]{article} 
\usepackage{aaai23}  
\usepackage{times}  
\usepackage{helvet}  
\usepackage{courier}  
\usepackage[hyphens]{url}  
\usepackage{graphicx} 
\usepackage{natbib}  
\usepackage{caption} 
\usepackage{tablefootnote}
\usepackage{algorithm}
\usepackage{algorithmic}
\usepackage{enumitem}
\usepackage{booktabs} 

\usepackage{color}
\usepackage{multirow}
\usepackage{stmaryrd}
\usepackage{trimclip}

\usepackage{graphicx,subcaption} 
\makeatletter
\DeclareRobustCommand{\shortto}{%
  \mathrel{\mathpalette\short@to\relax}%
}
\newcommand{\short@to}[2]{%
  \mkern2mu
  \clipbox{{.5\width} 0 0 0}{$\m@th#1\vphantom{+}{\shortrightarrow}$}%
  }

\usepackage{newfloat}
\usepackage{listings}
\DeclareCaptionStyle{ruled}{labelfont=normalfont,labelsep=colon,strut=off} 
\floatstyle{ruled}
\newfloat{listing}{tb}{lst}{}
\floatname{listing}{Listing}
\nocopyright 

\title{Tracking Progress in Multi-Agent Path Finding}

\author{
Bojie Shen,
Zhe Chen,
Muhammad Aamir Cheema\\
Daniel D. Harabor and
Peter J. Stuckey\\
}
\affiliations{
Faculty of Information Technology, Monash University, Melbourne, Australia\\
\{bojie.shen, zhe.chen, aamir.cheema, daniel.harabor, peter.stuckey\}@monash.edu
}

\usepackage{bibentry}

\begin{document}

\maketitle

\begin{abstract}
Multi-Agent Path Finding (MAPF) is an important core problem for many new and emerging industrial applications.
Many works appear on this topic each year, and a large number of substantial advancements and performance improvements have been
reported. 
Yet measuring overall progress in MAPF is difficult: 
there are many potential competitors, and the computational burden for comprehensive experimentation is prohibitively large. 
Moreover, detailed data from past experimentation is usually unavailable. 
In this work, we introduce a set of methodological and visualisation tools which can help the community establish clear indicators for 
state-of-the-art MAPF performance and which can facilitate large-scale comparisons between MAPF solvers. 
Our objectives are to lower the barrier of entry for new
researchers and to further promote the study of MAPF,  since progress in the area and the main challenges are made much clearer. 

\end{abstract}
\section{Introduction}
Multi-Agent Path Finding (MAPF) is a combinatorial problem that asks us to compute collision-free paths for teams of cooperative agents.  MAPF is a core problem in many important and emerging industry applications, such as automated warehousing \cite{warehouse}, video game team navigation \cite{spaceA_star}, and drone swarm coordination \cite{drone}. 
Perhaps due to its wide-ranging applicability, MAPF is studied in a
number of different communities, including Artificial Intelligence, Operations Research, and Robotics. 

In recent years, the number of publications on the topic of 
MAPF has exploded, as industrial interest continues to grow.
Many works now appear, across many different venues, 
and there have been substantial performance improvements. 
To track progress in the area, the community has developed a set of standardised MAPF benchmarks~\cite{SternSoCS19}, which cover a variety of popular application domains and synthetic/pathological test cases. In total there are more than 1.5 million standard instances with up to thousands of moving agents per instance. Unfortunately, the computational burden associated with running this benchmark is large, which means that most researchers attempt to solve only a limited subset of 
instances and then only compare against a limited subset of 
potential competitors. 
One of the problems with this approach is that it is not entirely clear where a given MAPF 
solver sits on the pareto-frontier that currently defines the 
state-of-the-art.
Another related problem is the availability of results data. 
Although published works include headline results, such as success rates and total problems solved, they typically do not mention which specific problems were solved, which were closed, and where the remaining gaps
are.
Supplementary data, such as concrete plans and best-known bounds, 
which can allow other researchers to verify claims and build on 
established results, are seldom available. 
Thus, despite notable advancements and readily available benchmark problem sets, we do not currently have a clear picture of overall progress in MAPF.

In this work, we introduce a new set of methodological and visualisation tools 
to facilitate comparisons between a wide range of MAPF methods. We also 
propose clear indicators that can help establish state-of-the-art performance.
We then undertake a large set of experiments, with several currently 
leading optimal and suboptimal solvers, in an attempt to map the current 
pareto-frontier. Finally, we propose a new online platform for
the MAPF community to track and validate further gains and to improve visibility for and access to existing solvers. 
We believe that these proposals can help identify the main strengths of existing research and the remaining challenges 
in the area. They can also be used to track progress on those challenges over time. 
Finally, we believe that these proposals can help to lower the barrier of entry for new research on the topic of MAPF.

\section{Background and Literature Review}

MAPF is a broad topic with many contributions from diverse perspectives. We briefly describe the best-known and 
arguably most studied variant called {\em classical MAPF}~\cite{SternSoCS19}.
In this problem, a team of cooperative agents moves across a 4-connected grid map. Agents are allowed to move from a current grid cell (or vertex) to another adjacent grid cell, or else wait at their current location. 
Each move or wait action has unit cost. Time is also discretised into unit-sized steps. We are asked to compute paths to move every agent, from its start position to its goal location. The paths need to be collision free, which means agents do not collide with static obstacles on the map nor with each other. Performance for classical MAPF is measured in a variety of ways. The most common metrics are the number of problems solved, the success rate (percentage of problems solved for a fixed number of agents), the plan cost (usually sum-of-costs or makespan), and the runtime. 


\subsection{Existing Benchmarks and Limitations}
Early MAPF research (and even some recent papers) typically conduct experiments over individually generated problem sets, e.g.,~\cite{spaceA_star, EPEA-star,MDD-SAT}.
Historically these problem sets have not been published as separate artefacts, which creates difficulties trying to reproduce experimental data,
even if reference implementations of the original works are made available.
In response to these challenges the MAPF community developed a standard benchmark suite~\cite{SternSoCS19}, to test performance of (classical) MAPF solvers across 33 maps from 6 different types of grid domains: (1) Game grids, originating from real video games; 
(2) Street maps, originating from layouts of real cities;
(3) Maze maps, synthetically generated and featuring different corridor sizes;
(4) Room maps, synthetic grids with open areas connected by narrow entrances;
(5) Open grids, with or without random obstacles; and
(6) Warehouse maps, synthetic maps imitating automated warehouse environments.

Each map has 25 random and 25 even scenarios. A random scenario includes randomly generated start and goal locations for up to 1000 instances (agents). 
On even scenarios, start and goal locations have an even distribution on distance, and the total number of location pairs (up to 7000) varies across different maps. 
There are more than 1.5 million location pairs in total, which means the same amount of problem instances.
Table~\ref{coverage} gives a complete summary.

The computational burden required to run the whole benchmark is large, which means 
researchers only compare proposed approaches against a small number of contemporaries and only on a limited subset of problem instances.
Table \ref{coverage} shows a summary of reported results for a recent set of leading optimal and suboptimal solvers: CBSH2-RCT~\cite{RTC2}, BCP~\cite{BCP}, ID-CBS~\cite{ID-CBS}, EECBS~\cite{EECBS}, FEECBS+~\cite{FEECBS}, PIBT2~\cite{[PIBT]}, LNS2~\cite{MAPF-LNS,MAPF-LNS2}, and LaCAM~\cite{Lacam}. 
Although most domains are considered, researchers typically select only 1 to 2 maps per domain for evaluation. The number of instances selected for experiments meanwhile is small relative to the total available instances. 
In particular, most research increases the number of selected start and goal locations by 5 or 10 until the selected algorithm returns a timeout failure. Notice that coverage across the benchmark (sub-)set is incomplete. 

Another related issue is that experimental results now appear in many different venues across the literature. Thus, it is increasingly difficult to keep track of all recent advancements. 
Moreover, although headline results are published (e.g., success rate and total problems solved), detailed supplementary material, to indicate exactly which problems were solved and how well, are usually not publicly accessible.
Short of large-scale re-evaluations of many published works, it has become difficult to know exactly how much progress is being made in the area, and it is difficult to know how to select an appropriate subset of competitors for experimental comparisons. 
As shown in Table \ref{coverage}, most research handles these issues by comparing against only 1 or 2 closest rivals, yet there are far more research works claiming state-of-the-art results each year.

\begin{table}[]
    \centering
    \small
    \begin{tabular}{c|c|c|c|c}
        \toprule
        \multirow{2}{*}{} & \multirow{2}{*}{Domains} & \multirow{2}{*}{Maps}& \multirow{2}{*}{Instances} & Compared \\
        &&&&algorithms\\
        \midrule
 \midrule
        CBSH2-RCT & 6 & 8 & 1,200 & 1 \\
        \hline
        BCP  & 5 & 16 & 4,430 & 2 \\
        \hline
        ID-CBS & 6 & 32 & 3,380 & 1 \\
        \hline
        EECBS & 4 & 6 & 1,200 & 3 \\
        \hline
        FEECBS+ & 6 & 8 & 1,800 & 1 \\
        \hline
        PIBT2 & 4 & 10 & 21,075 & 6 \\
        \hline
        LNS2 & 6 & 33 & 825 & 3\\
        \hline
        LaCAM & 5 & 12 & 5,000 & 6\\
        \hline
        Total Available & 6 & 33 & 1.5 M & ---  \\
    \bottomrule
    \bottomrule

    \end{tabular}
    \caption{The number of domains, maps, instances and compared algorithms in recent MAPF research.}
    \label{coverage}
\end{table}

\subsection{Existing Work on Large-scale Evaluation}

Several recent initiatives have attempted to track MAPF performance. 
The community website, \texttt{mapf.info}\footnote{http://mapf.info/index.php/Data/Data}, features a dedicated page that encourages researchers to report their results. 
But there is no consistent way to submit, manage, or explore these results. Thus, since its inception more than two years ago, only one algorithm has been reported here, with only summary results and no proof of claim.

Another recent attempt is due to \citet{eli-evaluation}, 
who conducted a large-scale experiment to compare
5 leading optimal MAPF algorithms. 
Authors evaluate 
all maps and scenarios from the MAPF benchmark suite.
Meanwhile in~\citet{ren2021mapfast}, authors evaluate six recent optimal algorithms, in another large-scale experiment. 
In these works, authors analyse the percentage of instances solved by each algorithm and observe general trends.
A main drawback of these research works is the limited coverage of algorithms, including notable exclusions such as BCP-7 \cite{BCP-7} and CBSH-RCT \cite{RCT} 
(two very successful and performant solvers). 
In addition, authors only give summary results.
For example, detailed progress for each domain and instance is unavailable. Plans, plan costs and best-known bounds are also unavailable, which makes it difficult to judge progress in the area, difficult to validate results and impossible to include new points of comparison (without re-producing the experiment).
Recently in~\citet{ewing2022betweenness} authors evaluate seven different algorithms to determine features that make MAPF problems hard. Although wide-ranging, experiments are on non-standard problem sets. Detailed data is unavailable.


\subsubsection{Discussion}
Substantial interest and large recent advances have grown the size of the MAPF community. 
These are positive developments, but they come at a price: progress is harder to track, the main challenges are less clear, and barriers for entry are increasing.
To continue growing the MAPF community needs effective tools: to track progress, investigate results, simplify comparisons and to help researchers stay up to date with recent developments.
%

\section{Methodology}


In this section, we introduce our methodology to track the progress of different methods on MAPF benchmarks. In general, there are three types of algorithms studied by the research community:

\begin{enumerate}[label=(\roman*)]
    \item \textbf{Optimal Algorithms} focus on finding exact optimal solutions. Such algorithms start from a lower-bound of the optimal solution, and 
    progressively push the lower-bound until they find a feasible solution that is provably optimal. 
    
    \item \textbf{Bounded Suboptimal Algorithms} find the suboptimal solution within theoretical guarantees. These algorithms explore the lower-bound 
    and feasible solutions simultaneously, and return the solution that is within certain suboptimality w.r.t. current best lower-bound.
    
    \item \textbf{Unbounded Suboptimal Algorithms} focus on finding feasible solutions. These algorithms find the feasible solution fast, and keep improving it given sufficient time.
\end{enumerate}

Our goal is to design a system that tracks different types of algorithms and their progress together. The critically important feature for us is the ability to handle all types of algorithms. 
Therefore, we focus on two important results reported by different MAPF algorithms: 
(a) best (i.e., largest) lower-bound value: we track this value to cover the algorithms in (i) and (ii); and (b) best (i.e., smallest SIC) solution: we record this result to cover the algorithms in (ii) and (iii).
In addition to bounds and costs, runtime is another frequently used metric that can further distinguish between competing solvers. 
We do not attempt to track this aspect owing to the large variability in configuration setups from one paper to the next. 
In the remainder of this section, we explain our strategies for generating insight and systemic analysis from results
data, as well as a list of important things that we are tracking on different levels of the benchmark.

\subsubsection{Instance-level Tracking}
At the instance level, our system records the best lower-bound and solution cost as explained above.
For each reported lower-bound or valid plan we also keep track of additional metadata, such as the algorithm which produced the result, names of authors, publication references and links to implementations. 
We then use the data to provide additional insights:
 
     \underline{\textit{Tracking the concrete plan: }} each instance contains a different number of agents, however, it is not clear how these agents are distributed w.r.t. the obstacles of map, and how their solution paths interact on the map. Our system records a concrete plan for each best known solution cost and provides a visualiser to better understand those solutions.
    
    \underline{\textit{Tracking the gap: } }
    For each instance, we may have different algorithms which contribute lower-bounds and solutions (upper bounds) separately. Together, we need to analyse how close these algorithms are in terms of finding and proving optimal solutions. Therefore, we automatically track and visualise the \textit{suboptimality ratio} of each instance defined as ($\mathcal{S} - \mathcal{L}$) / $\mathcal{L}$ where  $\mathcal{L}$ and  $\mathcal{S}$ are the best known lower-bound and solution of the instance, respectively.
  

\subsubsection{Scenario-level Tracking}
All instances in a scenario are categorised into three types: 
(i) \emph{closed instance}: the instance has the same best lower-bound and solution cost (indicating that the solution cannot be further improved);
(ii) \emph{solved instance}: the instance has a feasible solution reported, but the current best lower-bound is less than the solution cost (i.e., improvement may be possible);
and (iii) \emph{unknown instance}: the instance has no solution reported.
For each scenario, our system tracks the percentage of \emph{closed} and \emph{solved} instances to indicate the progress of all contributed algorithms. 
For scenarios of the same map, we also track the following:

    \underline{\textit{Tracking progress on scenarios: }}For a given map, our system automatically generates plots which shows the percentage of \emph{closed}, \emph{solved} and \emph{unknown} instances for every scenario. The objective here is to identify the scenarios that are hard to solve with existing MAPF algorithms, so that more attention can be paid to these.
    
    \underline{\textit{Tracking progress on different number of agents: }}
    Each scenario contains instances with different numbers of agents. It is important to understand the scalability of MAPF algorithms across all scenarios (i.e., at what number of agents we stop making progress). Therefore, our system includes the percentages of \emph{closed}, \emph{solved} and \emph{unknown} instances for different number of agents on the same map.

\subsubsection{Domain and Map-level Tracking}
Finally, at the map-level of the benchmark, our system records the percentages of \emph{closed} and \emph{solved} instances for each map. 
%
    Similar to the scenario-level, our system also generates plots to track the percentages of \emph{closed}, \emph{solved}, and \emph{unknown} instances across all maps, and summarises the related maps for each domain to provide domain-level plots. This allows researchers to focus their efforts on solving those parts of the benchmark that have seen only limited progress.



\subsubsection{Participation and Comparison}
Another critical feature of our system is allowing other researchers to participate by submitting their algorithms/results and establish the state-of-the-art together. 
For all the results we collect, we make them publicly available and allow other researchers to download the results at each level. 
In order to make it easy for researchers to evaluate their own progress against other attempts we also provide tools to automatically compare algorithms, across every level of the system. Our principal evaluation criteria are: \# of instances a given algorithm closed; \# of instances that the algorithm solved; \# of instances for which it achieved the best lower-bound; \# of instances for which it reported the best solution. We apply these four criteria to summarise the state-of-the-art for each type of algorithm (optimal, bounded- and unbounded-suboptimal).

\renewcommand{\thesubfigure}{\roman{subfigure}}
\begin{figure*}[t]
\scriptsize
\begin{tabular}{@{~}rrr@{~}}

\begin{minipage}{.33\linewidth}
\centering
\includegraphics[width=\linewidth]{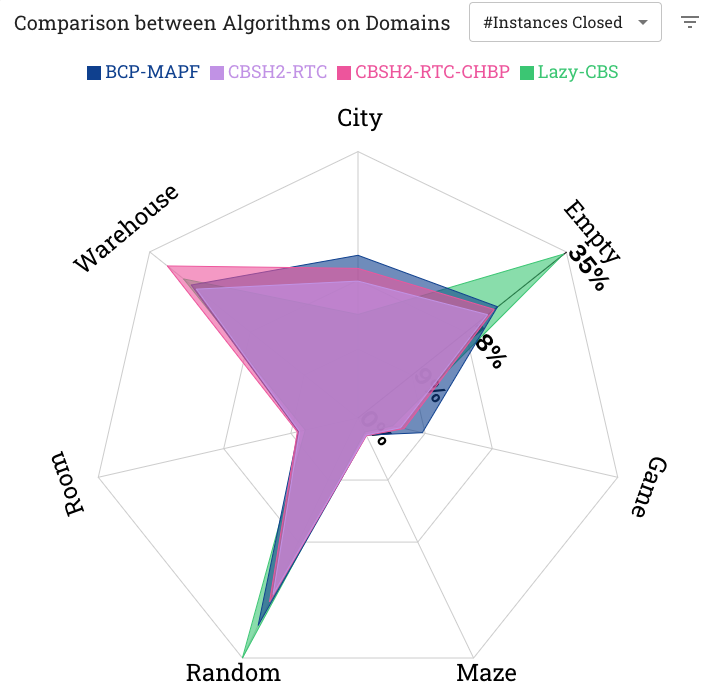}
\subcaption[]{}
\end{minipage}&

\begin{minipage}{.33\linewidth}
\centering
\includegraphics[width=\linewidth]{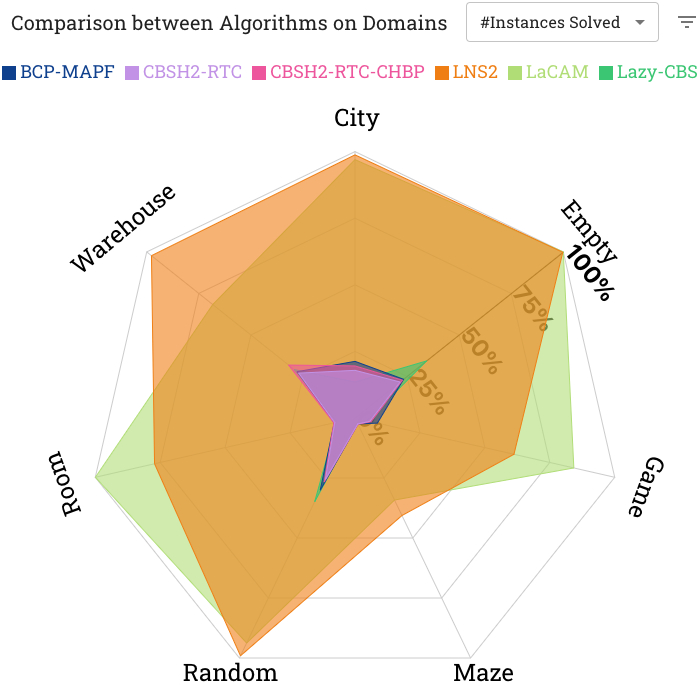}
\subcaption[]{}
\end{minipage} &

\begin{minipage}{.33\linewidth}
\centering
\includegraphics[width=\linewidth]{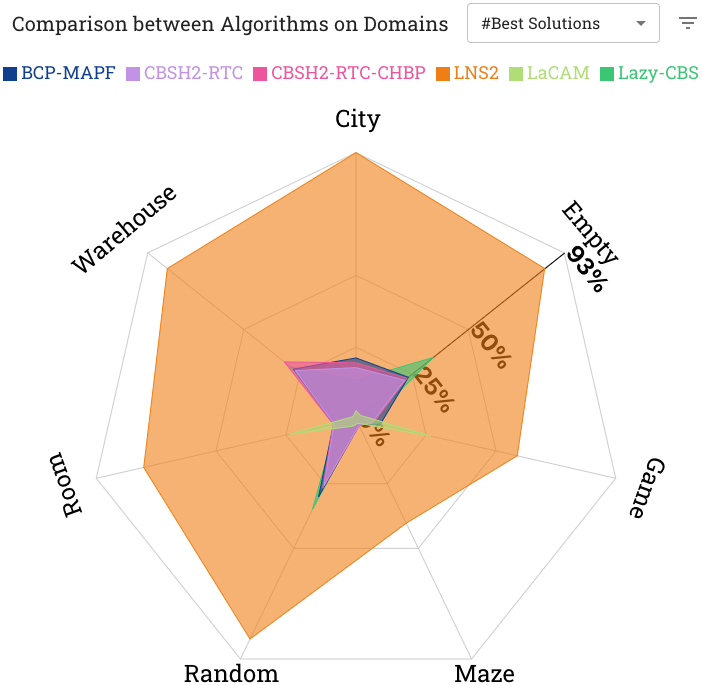}
\subcaption[]{}
\end{minipage}

\end{tabular}
\caption{Screenshots taken from our website. (i) Percentages of \# of instances closed; (ii) Percentages of \# of instances solved; and (iii) Percentages of \# of instances achieved best solution by each algorithm 
for various domains.
}
\label{fig::domain-level}
\end{figure*}

\subsubsection{Submission Interface}

The submission interface is meant to make it easy for anyone in the community to upload results for any of the benchmarks. A submission \emph{batch} requires a minimal set of metadata (details about the authors, the algorithm, its source code repository (if any)), and then consists of, for each tackled instance (i.e. map, scenario, number of agents) a lower bound (if the algorithm generates one), a solution cost (if the algorithm generates one), and a concrete plan that meets the solution cost. The plan format is an ASCII string that specifies the movements of agents at each timestep, e.g., the string "$udlrw$" represents a agent moving up, down, left, and right respectively, while $w$ represents waiting at its current location. 
This is a compact way of storing plans, and also extensible to more complex plan formats in the future. 
The format for the submission is a .csv file.
The system checks the validity of the plan and its solution cost as it processes each entry. Thus we can guarantee all upper bound information is valid.
We are unable to check lower bounds\footnote{Although a format for checkable proofs for lower bounds of MAPF problems would be a valuable resource for the community.} so they trusted by default. If any later stage we find a lower bound is violated by correct plan, then we have direct evidence that the lower bounding submission was erroneous. In this case, we remove all lower bounds in the batch from the system.

\section{Implementation and Initial Results}

We implement our proposed system as a website.\footnote{Our website is accessible at: \url{http://tracker.pathfinding.ai}. 
A demo video giving an overview of the system is also available at:   \url{http://tracker.pathfinding.ai/systemDemo}.
} 
For the back-end, we use Mongodb to manage all submissions in a database, and use Nodejs to implement APIs in order to communicate between database and web front-end.
For the front-end, we use a React interface.
To seed the database, we evaluate four different state-of-the-art optimal algorithms. This allows us to determine the best known lower bounds and optimal solutions. We also evaluate two leading unbounded-suboptimal algorithms, to explore the best known feasible solutions for as many as instances as possible. The details of the algorithms are shown in Table~\ref{algorithm_run}.
To ease the computational burden, we run each algorithm on each instance for one minute by default. For CBSH2-RCT and CBSH2-RCT-CHBP, we run the algorithm for one minute and check whether the lower-bound value is increased when finished. If so, we increase the time-out to another minute and keep searching. Otherwise, we terminate the algorithm. For all algorithms, we run every instance of a scenario by increasing the number of agents, and terminate if two instances in a row fail.

\setlength{\tabcolsep}{1.5pt}
\begin{table}[t]
    \centering
    \small
    \begin{tabular}{l|c}
    \toprule
        Algorithm Name & Type \\
 \midrule
 \midrule
        CBSH2-RCT~\tablefootnote{\url{https://github.com/Jiaoyang-Li/CBSH2-RTC}} ~\cite{RTC2}& Optimal\\
        \hline
        CBSH2-RCT-CHBP\tablefootnote{\url{https://github.com/bshen95/CBSH2-RTC-CHBP}}~\cite{CHBP}& Optimal \\
        
        \hline
        BCP-MAPF\tablefootnote{\url{https://github.com/ed-lam/bcp-mapf}}~\cite{BCP} & Optimal   \\
        \hline
        Lazy-CBS\tablefootnote{\url{https://bitbucket.org/gkgange/lazycbs}}~\cite{lazy-cbs}& Optimal  \\
        \hline
        LNS2\tablefootnote{\url{https://github.com/Jiaoyang-Li/MAPF-LNS2}} ~\cite{MAPF-LNS2} & Unb-suboptimal   \\
        \hline  
        LaCAM\tablefootnote{\url{https://github.com/Kei18/lacam}}~\cite{Lacam} & Unb-suboptimal   \\
  \bottomrule
\bottomrule


    \end{tabular}
    \caption{The Optimal and Unbounded Suboptimal (Unb-Suboptimal) algorithms run in our website.}
    \label{algorithm_run}
\end{table}

\begin{figure*}[t]
\scriptsize
\begin{tabular}{@{~}rrr@{~}}

\begin{minipage}{.48\linewidth}
\centering
\includegraphics[width=\linewidth]{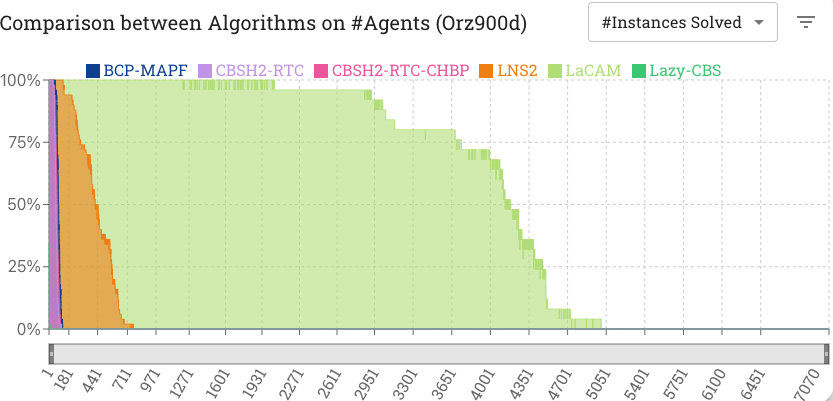}
\subcaption[]{}
\end{minipage}&

\begin{minipage}{.48\linewidth}
\centering
\includegraphics[width=\linewidth]{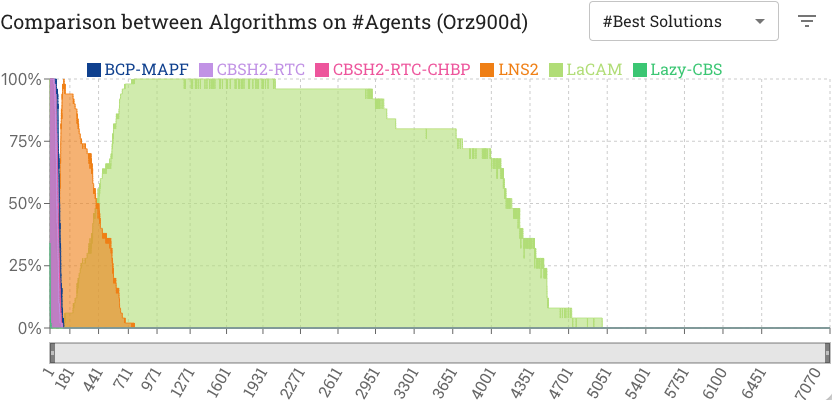}
\subcaption[]{}
\end{minipage} 

\end{tabular}
\caption{Screenshots taken from our website. (i) Percentages of instances solved; and (ii) Percentages of instances achieving best solution by each algorithm for different number of agents on x-axis, we show the result for game map: \texttt{orz900d}.
}
\label{fig::scen-level}
\end{figure*}

\begin{figure*}[t]
\scriptsize
\begin{tabular}{@{~}rr@{~}}

\begin{minipage}{.48\linewidth}
\centering
\includegraphics[width=\linewidth]{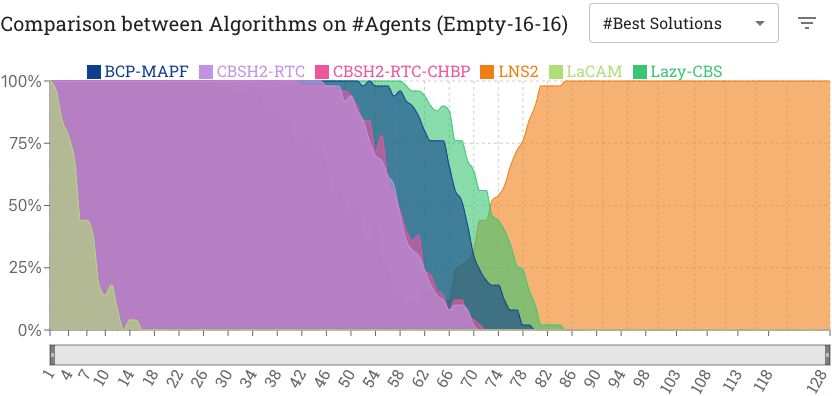}
\subcaption[]{}
\end{minipage}&

\begin{minipage}{.48\linewidth}
\centering
\includegraphics[width=\linewidth]{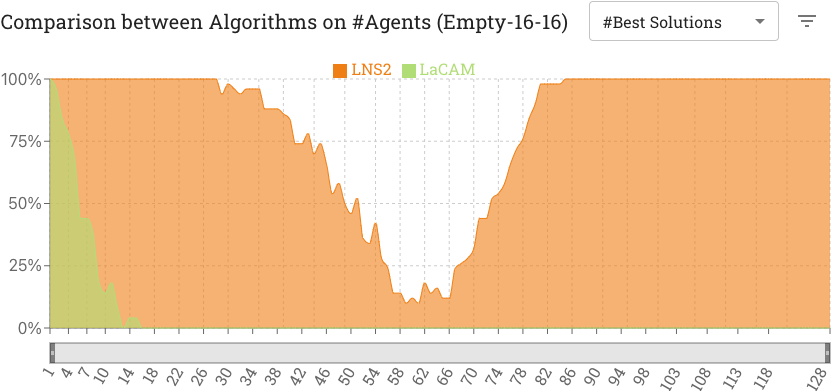}
\subcaption[]{}
\end{minipage} 

\end{tabular}
\caption{Screenshots taken from our website. 
Percentages of instances achieving best solution 
(i) by each algorithm; and (ii) by LNS2 and LACAM only, for different number of agents on x-axis, we show the result for game map: \texttt{empty-16-16}.
}
\label{fig::empty-scen-level}
\end{figure*}

\begin{figure}[t]
  \centering
  \includegraphics[width=\linewidth]{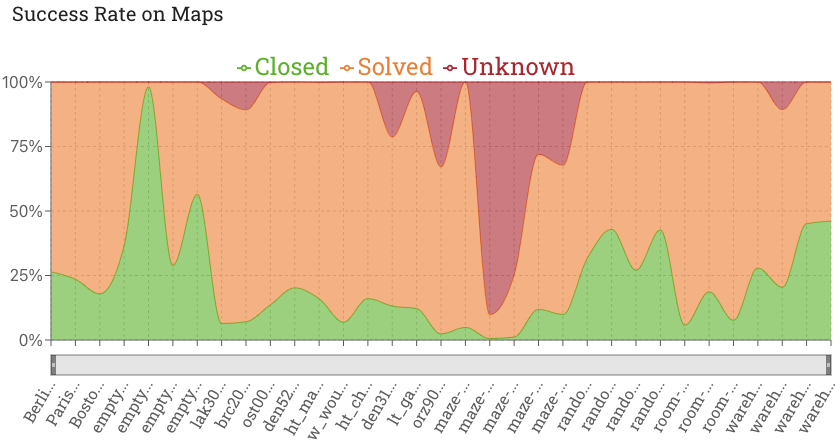}
    \caption{Screenshots taken from our website. Percentages of \# closed, solved and unknown instances shown for various maps on x-axis.}
  \label{fig::map_progress}
\end{figure}

\subsection{Domain and Map-level Analysis}

We demonstrate how our website can be used to explore the new pareto-frontier, as well as the insights for the submitted MAPF algorithms.
To begin, Figure~\ref{fig::domain-level} shows the plots that summarise the submitted algorithms on each domain. 

Figure~\ref{fig::domain-level} (i) shows the \# of instances closed by each algorithm (shown as percentage).
BCP-MAPF slightly outperforms the CBS variations, CBSH2-RTC and CBSH2-RTC-CHBP, on most of domains, but achieves a lower number of instances closed than Lazy-CBS on the Empty and Random maps. This shows that the exponential reduction in search that Lazy-CBS can achieve pays off on these smaller maps.
However, no existing solver is able to close many instances for domains such as Maze, Room and Games. Thus, more attention is needed on how to solve these domains optimally. 

Figure~\ref{fig::domain-level} (ii) shows the \# of instances solved by each algorithm. Clearly, 
LNS2 significantly outperforms the four optimal solvers in terms of \# of instance solved. LaCAM further mitigates the weaknesses of LNS2 and is able to solve more instances than LNS2 on domains such as Room and Game.
But, again, no solver is able effectively tackle maze maps.

Figure~\ref{fig::domain-level} (iii) shows the \# of best solutions achieved in order to compare the solution quality of the solved instances for each algorithm. Although LaCAM can solve more instances than LNS2 on Room and Game maps, surprisingly, the solution quality of the solved instances of LaCAM is dominated by LNS2 on every domain.  In fact, LaCAM is dominated by all other algorithms and finds the best solution when it is the only algorithm capable of finding a solution. 

To further dig into the details, 
Figure~\ref{fig::map_progress} demonstrates the collective progress that all algorithms have made for different maps in terms of the number of closed, solved and unknown instances.
Other than Maze maps, where no solver is able to perform well, almost all instances on all maps have been completely solved. 
Yet we are still a long way from being able to close all instances.
Additionally, we see that some of the large-scale game maps, such as \texttt{orz900d}, \texttt{den312d} and \texttt{brc202d}, are not only far from being completely solved yet, but they also have a low number of closed instances -- less than 15\% are provably optimal.
We now move our attention to the next level down, and focus on the map \texttt{orz900d}.

\begin{figure}[t]
  \centering
  \includegraphics[width=\linewidth]{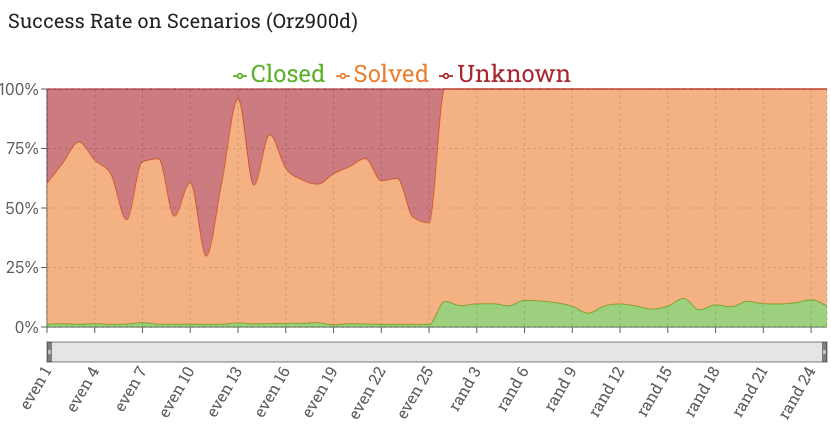}
    \caption{Screenshots taken from our website. Percentages of closed, solved and unknown instances shown for various scenarios of \texttt{orz900d} map on x-axis.}
  \label{fig::scen_progress}
\end{figure}

\subsection{Scenario-level Analysis}

Figure~\ref{fig::scen-level} shows the comparison between different solvers on different number of agents for the game map, \texttt{orz900d}.
On the left, Figure~\ref{fig::scen-level} (i) shows \# of instances solved by each algorithm, where we see the optimal solvers (e.g., BCP-MAPF, Lazy-CBS, CBSH2-RTC and CBSH2-RTC-CHBP) can only scale up to no more than 181 agents. LNS2 is able to solve the instances on more \# of agents than the optimal solvers with a maximum improvement of up to four times.
LaCAM can scale to many more agents 
Clearly, LaCAM outperforms the other solvers, the \# of instances it can solve safely scales to around two thousand agents and slowly decreases until reaching around five thousand agents.
Figure~\ref{fig::scen-level} (ii) presents the \# of instances where each algorithm achieved the best solution. Unsurprisingly, as we have seen before, LaCAM 
almost never achieves the best solutions on these instances that solved by LNS2 or any other solvers.

Note that Figure~\ref{fig::scen-level} (ii) does not really allow us to contrast the optimal solvers, since they all can only handle a few instances. Figure~\ref{fig::empty-scen-level} (i) show the same plots for the much smaller map \texttt{empty-16-16}, here we can see that CBSH2-RTC finds the best solution for about half the instances, and CBSH2-RTC-CHBP slightly improves upon this, BCP improves further, and Lazy-CBS even further. All of the optimal solvers fail after 85 agents. We can see that on some small instances where LaCAM also finds the optimal solution. 
LNS2 finds best solutions early-on (obscured in the plot) and then dips down, as it finds suboptimal solutions, then returns to 100\% (best known solution) for instances where the optimal solvers are unable to prove optimality. The platform allows us to just plot some solvers, restricting to LaCAM and LNS2, Figure~\ref{fig::empty-scen-level} (ii) shows how good LNS2 is at finding optimal solutions on smaller instances. This illustrates how we can use the platform to easily compare solvers on a single map. We now return to analysing \texttt{orz900d}.


Figure~\ref{fig::scen_progress} shows the collective progress that all algorithms have made across the different scenarios of \texttt{orz900d}, again in terms of the number of closed, solved and unknown instances.
Interestingly, although we have seen there is a large proportion of instances 
not solved when the \# of agents is high, all randomly generated scenarios (i.e., rand 1 - 25) are 100\% solved, the unsolved instances only arise in the evenly generated scenarios (i.e., even 1 - 25). This is because the randomly generated scenarios contain 
up to one thousand agents, where as the evenly generated scenarios have around six thousand agents in each scenario file.
In even scenarios, the start and goal locations are generated based on the maximum grid distance, $d_{max}$, between any two traversable cells on the map. The map is divided into $\lfloor d_{max}/4 + 1 \rfloor$ buckets of $(s,g)$ pairs, where the $i$-th bucket contains 10 $(s,g)$ pairs with a grid distance between them within the range of $i \times 4$ to $(i+1) \times 4$. Thus, the number of agents can become large as the size of the map increases.
Next, we zoom in to the next-level, to further explore the insight of even-1 scenarios.


\begin{figure}[t]
  \centering
  \includegraphics[width=\linewidth]{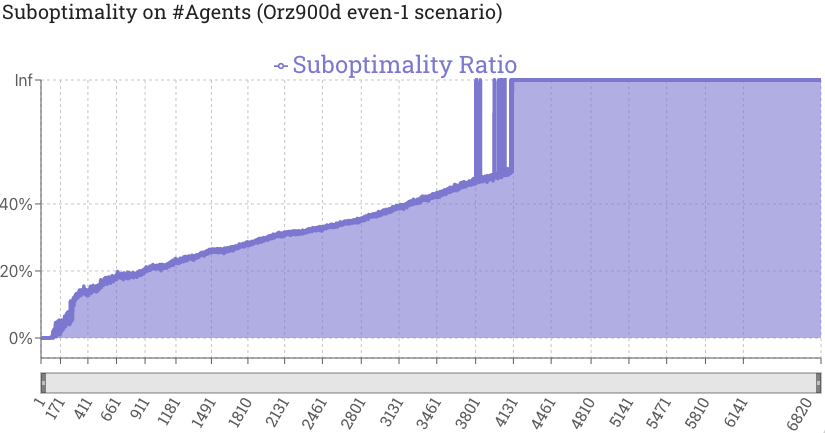}
    \caption{Screenshots taken from our website. The suboptimality ratio (i.e., ($\mathcal{S} - \mathcal{L}$) / $\mathcal{L}$ where  $\mathcal{L}$ and  $\mathcal{S}$ are the best known lower-bound and solution of the instance) on different \# agents for even-1 scenario on \texttt{orz900d} map.  }
  \label{fig::instance_progress}
\end{figure}

\begin{figure}[t]
  \centering
  \includegraphics[width=\linewidth]{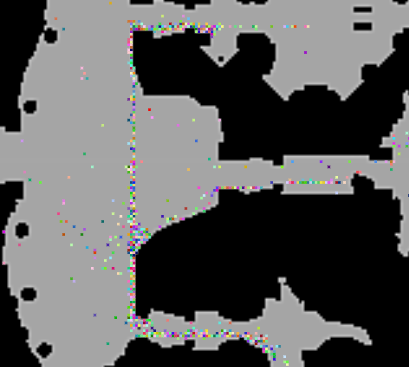}
    \caption{
    A feasible solution has been found for an even-1 scenario instance on the \texttt{orz900d} map, where \# of agents is 4119.
    We show the screenshot on part of the map from the visualizer of our website. The colored points represent individual agents.}
  \label{fig::instance_example}
\end{figure}

\subsection{Instance-level Analysis}
Figure~\ref{fig::instance_progress} shows 
the suboptimality ratio between best known lower-bound and solution of instances with various \# agents on x-axis. 
Since most instances for the even-1 scenario of \texttt{orz900d} are solved only by LaCAM, the suboptimality is computed based on the trivial lower bound (i.e., the SIC of each agent follows the shortest path by ignoring other agents), thus the value indicates the upper bound of suboptimality, where the solution quality of solved instances for LaCAM is no more than around 40\% worse than the optimal solution.

Now, let us focus on a particular instance in order to analyse why other algorithms can not solve it. 
Figure~\ref{fig::instance_example} shows a screenshot of a feasible solution for an instance of even-1 scenario on part of the \texttt{orz900d} map, where the \# agents is 4119. 
From the figure we see that substantial congestion forms around obstacle corners.  Tightly bending around corners is a necessary condition for individually optimal paths, but with many agents on the map, following those individually optimal paths incurs substantial delays. We see that agents wait and must form queues to reach their goal locations. This behaviour is characteristic of the Push-and-Swap~\cite{[PIBT],push_swap} strategy used in LaCAM. This strategy essentially asks each agent to follow its shortest path if possible, but tries to push or swap the locations with other agents when they are in conflict. Optimal solvers can not solve this instance, because they rely on complicated reasoning techniques to resolve the conflicts occurring in an infeasible plan. Indeed such techniques can only scale up to hundreds of agents.
LNS2 meanwhile uses a two-step framework that first plans an initial solution using prioritised planning and then improves that solution using a Large Neighbourhood Search (LNS). The core of this approach -- prioritised planning -- is known for being very fast and effective, especially on large maps where many paths exist for each agent. Why then is LNS2 unable to solve this instance?

In this case, our visualisation shows that planning 
in the congested area requires avoiding many 
temporal obstacles. We also see that agent start locations are far from their target positions. The combination (long path, many temporal obstacles) 
explodes the size of the time domain. The effect 
is that individual path planning times increase, from milliseconds per query on other maps to several seconds per query on this map. 
We now see that LNS2 fails because there is not enough time to even compute an initial plan. 
%
This analysis points out a place where researchers 
might try to improve the state-of-the-art: by developing better and more effective low-level solvers for the LNS2 algorithm. It seems likely that such cases were not considered during the original development.
Fortunately, LNS2 can work with any initial solution, just using its second phase to improve it. 
One notable takeaway from our analysis is the potential for combining LNS2 and LaCAM. 
In the combination, we can leverage LaCAM to obtain an initial solution quickly, while utilising LNS to continuously improve the solution quality in remaining computation time. This combination effectively addresses the drawbacks of both algorithms at the same time, and represents an interesting direction for future work.

\section{Conclusion and Future Work}
We introduce a new set of methodologies to track the progress of MAPF from various perspectives.
We evaluate several currently leading optimal and suboptimal solvers on a large-scale experiment.
The detailed results are publicly accessible and illustrated by a set of well-built visualisation tools on our online platform.
We believe that our platform helps: identify the remaining challenges in MAPF problems; establish state-of-the-art performance;
and provide a valuable dynamic resource for understanding the various challenges arising in MAPF.

Currently, the platform 
concentrates only on lower and upper bounds for problem instances. 
This is usual in the world of optimisation problems, since it allows us to understand the gap between lower and upper bounds, and hence how far away an instance is likely to be from being \emph{closed}. It also allows us to attribute the closure of instances to particular algorithms. The platform does not consider run times, and indeed many of the results currently there could be improved by giving more runtime to the algorithms.\footnote{Although in many cases doubling or even multiplying available time by 10 will make little difference, and CBS algorithms may run out of memory before reaching the time limit~\cite{ID-CBS}.}
Clearly runtimes are of interest in practice. 
But runtime is hard to compare fairly (over different machines), and indeed there is no way to guarantee claimed runtimes as part of a submission are indeed correct.
A future direction for the platform is to allow executables to be submitted, and then run for a fixed amount of wall clock time on each instance. This would give us a more detailed view of the state-of-the-art from the run-time perspective. Another interesting direction is to extend the platform to different types of agent models. This would allow the community to track progress on a variety of different MAPF variants, which all share
common benchmarks; e.g., Continuous-time MAPF~\cite{continue_mapf}.

\section{Acknowledgements}

This work is supported by the Australian Research Council under grant DP200100025, DP230100081, FT180100140, and by a gift from Amazon.


\bibliography{aaai23}
\bibstyle{aaai23}

\end{document}